\pdfoutput=1
\documentclass[11pt]{article}

\usepackage[final]{acl}

\usepackage{times}
\usepackage{latexsym}

\usepackage[T1]{fontenc}
\usepackage[utf8]{inputenc}

\usepackage{microtype}

\usepackage{inconsolata}

\usepackage{graphicx}

\usepackage{multirow}
\def\CHK#1 {\textcolor{magenta}{{\bf [CHK:}~#1{\bf ]}}~}
\def\ADD#1 {\textcolor{cyan}{{\bf [ADD:}~#1{\bf}]}~}

\usepackage{amssymb} % enable /mathbb
\usepackage{amsmath} % enable /mathbb
\usepackage{tcolorbox}
\usepackage{multicol}

\usepackage{booktabs}

\title{HAIC: Improving Human Action Understanding and Generation with Better Captions for Multi-modal Large Language Models}

\author{
\textbf{Xiao Wang}\textnormal{\textsuperscript{1}\footnotemark[1]\footnotemark[2]} \quad
\textbf{Jingyun Hua}\textnormal{\textsuperscript{1}\footnotemark[1]} \quad
\textbf{Weihong Lin}\textnormal{\textsuperscript{1}\footnotemark[1]} \quad
\textbf{Yuanxing Zhang}\textnormal{\textsuperscript{1}} \quad \\
\textbf{Fuzheng Zhang}\textsuperscript{1} \quad
\textbf{Jianlong Wu}\textsuperscript{2}\footnotemark[3] \quad
\textbf{Di Zhang}\textsuperscript{1} \quad
\textbf{Liqiang Nie}\textsuperscript{2}\footnotemark[3] \\
\textsuperscript{1}Kuaishou Technology \quad
\textsuperscript{2}Shandong University \\
{\tt\small scz.wangxiao@gmail.com, huajingyun@hotmail.com, lwher1996@outlook.com, longo@pku.edu.cn} \\
{\tt\small zhfzhkris@outlook.com, jlwu1992@pku.edu.cn, zhangdi08@kuaishou.com, nieliqiang@gmail.com}
}

\begin{document}
\maketitle

\renewcommand{\thefootnote}{\fnsymbol{footnote}}  % 让脚注符号变为符号模式
\footnotetext[1]{Equal contribution.}  
\footnotetext[2]{Work done as an intern at Kuaishou Technology.}
\footnotetext[3]{Corresponding author.}

\begin{abstract}

Recent Multi-modal Large Language Models (MLLMs) have made great progress in video understanding. However, their performance on videos involving human actions is still limited by the lack of high-quality data. 
To address this, we introduce a two-stage data annotation pipeline. First, we design strategies to accumulate videos featuring clear human actions from the Internet. 
Second, videos are annotated in a standardized caption format that uses human attributes to distinguish individuals and chronologically details their actions and interactions.
Through this pipeline, we curate two datasets, namely HAICTrain and HAICBench. \textbf{HAICTrain} comprises 126K video-caption pairs generated by Gemini-Pro and verified for training purposes. Meanwhile, \textbf{HAICBench} includes 412 manually annotated video-caption pairs and 2,000 QA pairs, for a comprehensive evaluation of human action understanding.
Experimental results demonstrate that training with HAICTrain not only significantly enhances human understanding abilities across 4 benchmarks, but can also improve text-to-video generation results.
Both the HAICTrain and HAICBench are released at \url{https://huggingface.co/datasets/KuaishouHAIC/HAIC}.

% ActionHub
\end{abstract}

% 整体上看贡献比较小，只有数据过滤流程。

% 怎么强化模型的贡献：提出模型的贡献。

% 

\section{Introduction}

% 卖点： 密集全面的动作caption、多主体、细粒度动作

Multi-modal large language models have notably showcased their preeminence across various video understanding tasks \cite{chen_internvl2_2024, li_llava-onevision_2024, wang_qwen2vl_2024, zhang_llava-video_2024}. Among these, human action understanding plays a critical role in many downstream applications, e.g., human-computer interaction \cite{hayes2011hci}, autonomous driving \cite{xu2021autonomous}, embodied intelligence \cite{gupta2021embodied}, and human video generation \cite{wang2024koala}.

A recent study, ShareGPT4Video \cite{chen2024sharegpt4video} has demonstrated that high-quality and detailed video captions can improve MLLMs' performance in video understanding. However, most existing works \cite{ucf101_2012, xu_msr-vtt_2016, krishna_activitynet-caption_2017, wang_vatex_2019, chen_panda_70m_2024} provide only coarse captions for human actions, insufficient for understanding fine-grained behaviors. 
MotionLLM \cite{chen_motionllm_2024} introduces the MoVid dataset with fine-grained action captions from MotionX \cite{lin2023motionx}. Nevertheless, this dataset mainly focused on single-person scenarios. For multi-person situations, MoVid only considered consistent group activities like ``a group of people performing the Korean dance''. 
A comprehensive dataset is essential to enhance MLLMs' understanding of detailed human actions and interactions in both single- and multi-person contexts, critical for tasks like emotional analysis, motivation prediction, and relationship modeling.

\begin{figure*}[t]
    \includegraphics[width=\linewidth]{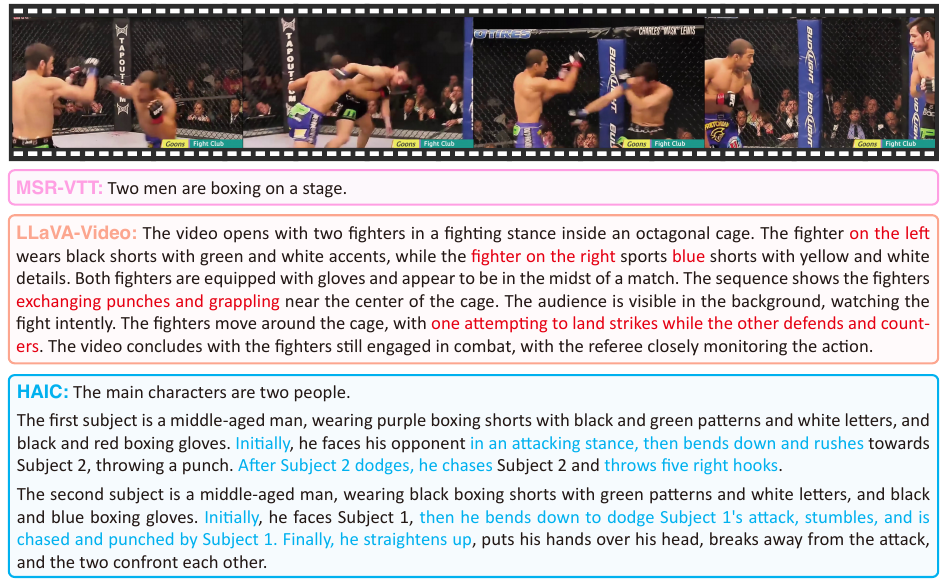}
    \caption{Our standardized caption format presents each individual's detailed attributes, body actions, and interactions in chronological order, making it easier to distinguish individuals and comprehend their behaviors.}
    \label{fig:intro}
\end{figure*}

There are two main challenges for building such datasets: (1) \textit{Action Video Accumulating.} How to automatically accumulate large-scale videos featuring clear actions of multiple individuals. (2) \textit{Caption formatting.} How to define a caption format that can clearly distinguish different people and detail their behaviors and interactions respectively.

To address the above challenges, we propose a novel data generation pipeline composed of two stages. 
In the \textit{video accumulation stage}, we accumulate videos from various domains featuring clear, meaningful human actions and identify their specific timestamps. This process is highly selective, with only about 1\% of videos meeting our quality criteria after applying various strategies.
In the \textit{caption annotation stage}, we define a caption format that uses human attributes to distinguish individuals and chronologically annotates detailed body actions and interactions for each person (see \autoref{fig:intro}). 
With this pipeline, we curate two datasets: \textbf{HAICTrain} (\textbf{H}uman \textbf{A}ction and \textbf{I}nteraction \textbf{C}omprehension \textbf{Train}ing set) and \textbf{HAICBench} (a benchmark for evaluation). 
HAICTrain contains 126K videos accumulated from WebVid \cite{bain_webvid_2021}, annotated in our defined caption format by Gemini-1.5-Pro \cite{team2024gemini}. 
HAICBench includes 412 YouTube videos with human-annotated captions in the same format. Furthermore, we generate 2,000 multiple-choice QA pairs across five categories---human interaction, action detail, action sequence, count, and human attribute---by prompting GPT-4o \cite{hurst2024gpt4o} and Gemini-1.5-Pro. Note that all the above machine annotation results undergo review and refinement by human annotators. 
Experimental results indicate that utilizing HAICTrain for training can remarkably enhance the model's human action understanding ability by 1\%-2\%. Additionally, in MovieGenBench \cite{polyak2024moviegenbench}, our post-trained model surpasses the original model in GSB score of 2.15 and 6.81 in HunyuanVideo \cite{hunyuan_video} and Wanx2.1 \cite{tongyi_wanxiang_2023}, respectively.

\begin{table*}[t]
% \small
\centering

\resizebox{\textwidth}{!}{

\begin{tabular}{cl}
\toprule[1pt]

\textbf{Category}           & \multicolumn{1}{c}{{\color[HTML]{1F2329} \textbf{Example}}}                                                           \\ \toprule[1pt]
                            & {\color[HTML]{016DCB} \textit{What gesture does the middle-aged woman make while talking to the other man?}}          \\
\multirow{-2}{*}{Interaction} &
  {\color[HTML]{1F2329} \begin{tabular}[c]{@{}l@{}}(A) Gestures with both hands clasped in front of her\\ (B) Claps her hands (C) Waves her hands in the air (D) Points at the desk\end{tabular}} \\ \hline
                            & {\color[HTML]{016DCB} \textit{What does the man in the black hat do with his right hand as he starts skateboarding?}} \\
\multirow{-2}{*}{Action Details} &
  {\color[HTML]{1F2329} \begin{tabular}[c]{@{}l@{}}(A) He waves it in greeting (B) He points down the slope \\ (C) He places it on his hip (D) He keeps it in his pocket\end{tabular}} \\ \hline
                            & {\color[HTML]{016DCB} \textit{What does the man in the gray cap do immediately after gripping the barbell?}}          \\
\multirow{-2}{*}{Action Sequence} &
  {\color[HTML]{1F2329} \begin{tabular}[c]{@{}l@{}}(A) He looks at the camera (B) He adjusts his hat (C) He bends down and lowers the barbell\\ (D) He walks towards his front left and rubs his hands together\end{tabular}} \\ \hline
                            & {\color[HTML]{016DCB} \textit{How many times does the man clap his hands?}}                                           \\
\multirow{-2}{*}{Count}     & {\color[HTML]{1F2329} (A) Three (B) Four times (C) Once (D) Twice}                                                    \\ \hline
                            & {\color[HTML]{016DCB} \textit{What color is the cropped jacket worn by the young female character in the video?}}     \\
\multirow{-2}{*}{Attribute} & {\color[HTML]{1F2329} (A) Pink (B) White (C) Blue (D) Black}                                                          \\
\bottomrule[1pt]
\end{tabular}

}

\caption{Task examples of HAICBench, showcasing a comprehensive human action understanding across spatial (action details), temporal (sequence, count), and multi-human interaction (interaction and attribute) aspects.
}
\label{tab:task_example}

\end{table*}

% @王霄 我建议第一个contribution改成大概这样，主要是要说一下，我们不是对着评测集造了点数据，而是主动造了数据对快速训练收敛和提升模型理解有帮助。
% We propose a novel annotation pipeline to provide data that can facilitate the motion understanding of the MLLMs:1) accumulate large-scale videos featuring clear actions with single or multiple individuals from the Internet, and 2) generate captions in a standardized format that can clearly distinguish different individuals and detail their actions and interactions.

Our contributions can be summarized as follows:
\begin{itemize}
    \item We propose a novel data annotation pipeline to provide data that can facilitate human action understanding, which 1) accumulates large-scale videos with clear actions from the Internet and 2) generates standardized captions that distinguish individuals and detail their actions and interactions.
    \item We introduce two datasets: HAICTrain which includes 126K generated-then-verified high-quality video-caption pairs for training; and HAICBench comprising 412 human annotated video-caption pairs and 2,000 QA pairs, designed to evaluate MLLMs' human action understanding comprehensively.
    \item Experiments demonstrate that training with HAICTrain significantly improves human action understanding in benchmarks including MVBench \cite{li_mvbench_2024}, PerceptionTest \cite{patraucean2024perception},  ActivityNet-QA \cite{yu_activitynet-qa_2019} and HAICBench. 
    Furthermore, HAICTrain substantially improves text-to-video generation on MovieGenBench.
\end{itemize}

\begin{figure*}[t] 
    \includegraphics[width=\linewidth]{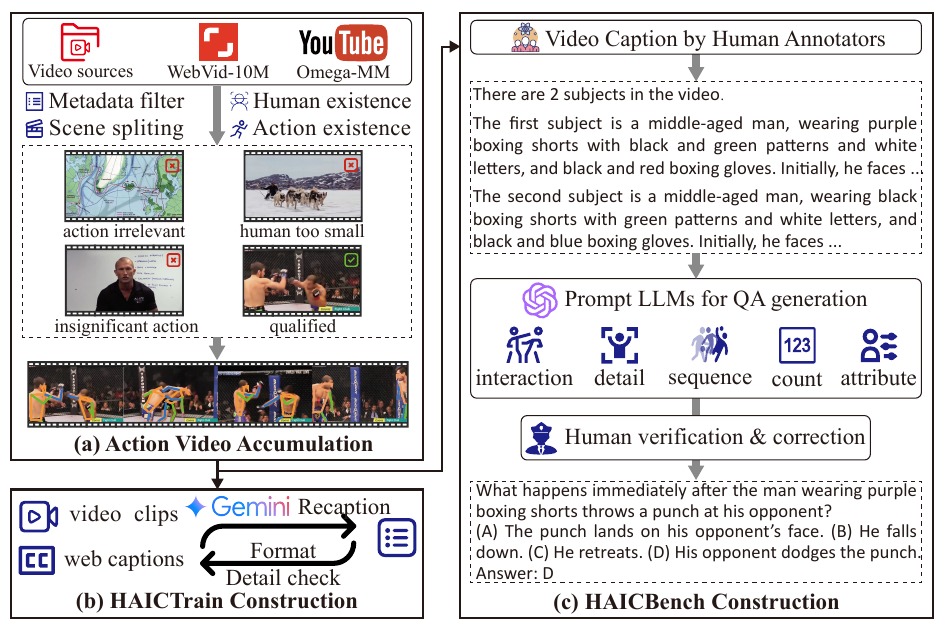}
    \caption {Our data generation pipeline. (a) The video accumulation stage collects videos featuring clear human actions from the Internet. Based on this, (b) HAICTrain is curated through Gemini-1.5-Pro re-captioning, and
    (c) HAICBench is created by LLM-assisted human annotation.
    }
    \label{fig:pipeline}
\end{figure*}

\section{Related Work}

\subsection{Video Caption Datasets}

Most existing video captioning datasets prioritized general video understanding \cite{xu_msr-vtt_2016, youcook_2016, wang_vatex_2019, miech_howto100m_2019, bain_webvid_2021, zellers_yt-tmp-180m_2021, qvhighlights_2021, xue_hd-vila_2022, chen2024sharegpt4video, chen_panda_70m_2024, wang_internvid_2024, wang_vid_dataflywheel_2024, xiong_lvd_2024, wang2024koala}. These datasets included only a subset of human videos, and the action captions tend to be coarse.
Some datasets specifically focused on human-centric captioning, including ActivityNet-Captions \cite{krishna_activitynet-caption_2017}, Ego4D \cite{ego4d_2022}, Kinetic-GEB \cite{Kinetic_GEB_222}, ActionHub \cite{actionhub_2024} and OpenHumanVid \cite{li_openhumanvid_2024}. 

Different from these datasets, our work focuses on finer-grained human actions and interaction understanding. 
% For example, videos of people talking are excluded from HAICTrain but included in OpenHumanVid.

\subsection{Video Understanding Benchmarks}

Traditional video understanding benchmarks have primarily honed in on a fixed set of classes \cite{ucf101_2012, amir_ntu_rgbd_2016, carreira_kinetics_2017, goyal_ssv2_2017}.
Recently, there has been a shift towards open-set video understanding, including video captioning and question answering \cite{xu_msr-vtt_2016, wang_vatex_2019, xiao_next-qa_2021, li_mvbench_2024}. While benchmarks like NeXT-QA \cite{xiao_next-qa_2021}, MVBench \cite{li_mvbench_2024}, and PerceptionTest \cite{patraucean_perception-test_2023} include action-related questions, they primarily focus on general action recognition rather than fine-grained details.
TemporalBench \cite{cai_temporalbench_2024} evaluates event-level action understanding using a different protocol that requires distinguishing correct captions from negative ones.
Action-specific benchmarks, such as ActivityNet-Captions \cite{krishna_activitynet-caption_2017}, TGIF-QA \cite{jang_tgif-qa_2019}, MoVid-Bench \cite{chen_motionllm_2024}, EGOBody \cite{zhang2022egobody}, and GRAB \cite{taheri2020grab}, address temporal sequences or human-object interactions but often overlook human-human interactions.

Our HAICBench provides a more comprehensive evaluation of human action understanding, covering five key aspects detailed in \autoref{tab:task_example}. In \autoref{sec_appe:comp_con_benchs}, we compare HAICBench with three concurrent works MoVidBench~\cite{chen2024motionllm}, MotionBench~\cite{hong2025motionbench}, and FAVORBench~\cite{tu2025favor}.

\subsection{Multi-modal Large Language Models} \label{sec:related_mllms}

MLLMs~\cite{liu_llava_2023, wang2024retake, wang2025adaretake} generally process multi-modal input information and generate language output. Kosmos \cite{huang_cosmos_2023} introduced an end-to-end framework that integrated visual inputs with LLM from cohesive training. Flamingo \cite{alayrac_flamingo_2022} and InstructBLIP \cite{dai_instructblip_2023} merged visual and linguistic features through cross-attention and a Q-Former module, respectively. MiniGPT-4 \cite{zhu_minigpt-4_2024} and LLaVA \cite{liu_llava_2023} simplified the integration by linearly projecting visual features directly into the LLM embedding space.

Recent studies focused on different aspects to enhance the above early attempts in MLLMs. 
Cosmos-2 \cite{peng_cosmos2_2024} and NeXT-GPT \cite{wu_next-gpt_2024} have expanded MLLM applications to broader multi-modal tasks.
LLaVA-1.5 \cite{liu_llava15_2024} explored adding high-quality multi-task training data, and scaling up the resolution and LLM size to boost MLLM performance.
LLaVA-OneVision \cite{li_llava-onevision_2024} explored to unify dynamic image resolution, multi-image, and video into a unified input format.
%
% Cambrian-1 \cite{tong_cambrian-1_2024} combined features from multiple complementary vision encoders with a Spatial Vision Aggregator for a more capable MLLM.

% All above MLLMs share a similar architecture, comprising vision foundational models, LLMs, and their connectors. We chose LLaVA-Video \cite{zhang_llava-video_2024} as the baseline. This model adopts a multi-layer perception as the connector which introduces minimal inductive bias, and leverages high-quality training data to achieve cutting-edge performance.

\section{HAIC Data Pipeline}

In this section, we detail our data generation pipeline, as illustrated in \autoref{fig:pipeline}.

\subsection{Action Video Accumulation} \label{sec:action_video_accu}

This pipeline aims to accumulate human videos featuring clear human actions and sufficient details from large-scale videos.

\noindent \textbf{Metadata Filtering.} We begin by discarding low-resolution videos and those without verbs in their descriptions using spaCy \cite{Honnibal_spaCy_Industrial-strength_Natural_2020}. 
The remaining videos are split into short clips with unique scenes using SceneDetect \cite{Castellano_PySceneDetect}. We keep clips between 5 and 20s long, as actions typically occur within a single scene.

\noindent \textbf{Human Existence Filtering.} We uniformly sample 16 frames from each clip and use the RTMPose \cite{jiang_rtmpose_2023} object detector to identify humans. Only videos where all frames contain 1-5 humans and the total bounding box area covers at least 10\% of the frames are retained, ensuring sufficient human detail.

\noindent \textbf{Human Action Filtering.} Using RTMPose \cite{jiang_rtmpose_2023}, we detect human bounding boxes and 17 body keypoints at 1 fps. Tracklets are constructed based on the maximal IoU between frames. We then filter out videos with static humans by ensuring the $L_1$ distance between all adjacent keypoints exceeds 0.085, with keypoint coordinates normalized by the video resolution. This filtering captures clear human actions.

However, we observe that 15\% of the filtered videos still contain static humans despite large keypoint $L_1$ distances. This often results from camera movements or image gallery videos. In these cases, we empirically find keypoints approximately follow an affine transformation (e.g., translation, scaling, rotation, and shear mapping). Based on this insight, we developed a strategy to filter these videos.
Formally, let the keypoint vector in frame $t$ be $\mathbf{P}_t\in\mathbb{R}^{3\times 17}$, where $17$ is the number of keypoints and $3$ corresponds to homogeneous coordinates $(\textmd{height}, \textmd{width}, 1)$. We assume keypoints in these videos generally adhere to the affine transformation:
\begin{equation} \label{eq:affine_trans}
    \begin{cases}
    \mathbf{P}_{t+1} = \mathbf{T} \mathbf{P}_t, \\
    \mathbf{T} =
    \begin{bmatrix}
         \mathbf{A} & \mathbf{t} \\
         \mathbf{O} & 1
    \end{bmatrix},
    \end{cases}
\end{equation}
where $\mathbf{A}\in\mathbb{R}^{2\times 2}$, $\mathbf{t}\in\mathbb{R}^{2\times 1}$ is the transformation coefficients, and $\mathbf{O}=[0,0]$. We solve the following least squares problems:
\begin{equation}
    \min_{\mathbf{A}, \mathbf{t}} { \left \|  
        \mathbf{P}_{t+1} - \mathbf{T} \mathbf{P}_t
    \right \|_2 },
\end{equation}
and retain only those samples with a residual value $r>0.0016$. A larger residual indicates a greater deviation from \autoref{eq:affine_trans}, suggesting that the unwanted videos mentioned above.

Overall, the whole action video accumulation step yields 0.31\% to 1.3\% of human action videos, depending on the video source.

\subsection{HAICTrain} \label{sec:fluid_cap}
We chose the WebVid-10M dataset \cite{bain_webvid_2021} as the video source for training due to its large scale and high vision quality. Initially, we apply the action video accumulation pipeline in \autoref{sec:action_video_accu} to collect action videos from the WebVid-10M. This process results in a collection of 126K videos, representing 1.2\% of the original dataset. Then, we employ Gemini-1.5-Pro \cite{team2024gemini} to generate captions in the standardized format as depicted in Figure \ref{fig:intro} referring to the videos and original captions. The specific prompts used for this process are detailed in \autoref{sec_appe:cap_generation}. We then employ an additional judgement to filter out failure cases that do not follow the pre-defined format or display low quality. This judgement ensures the quality of our data in HAICTrain.

\begin{figure*}[t] 
    \centering
    \includegraphics[width=0.7\linewidth]{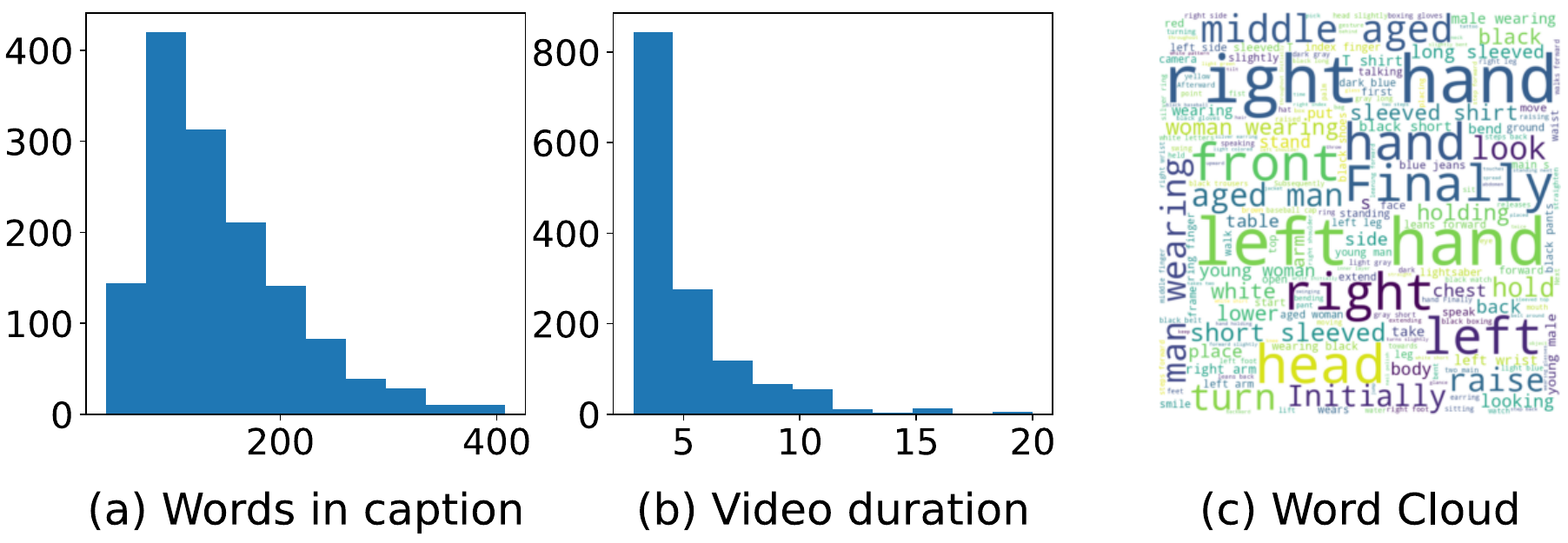}
    \caption {Statistics of HAICBench. Although videos are relatively short, the video captions are of high details including various action details and sequential actions.
    }
    \label{fig:bench_stats}
\end{figure*}

\subsection{HAICBench} \label{sec:fluid_bench}

We develop an LLM-assisted human annotation pipeline to create the HAICBench, which evaluates the capabilities of MLLMs in human action understanding. Human annotators first craft video captions following the format in \autoref{fig:intro}. To enhance question diversity, we then adopt LLMs to generate QA pairs based on these captions, which are then verified by annotators.

\noindent \textbf{Video Caption Annotation.} To avoid potential overlap with public benchmarks, we choose the newly proposed Omega-multimodal dataset \cite{omega_2024} as our benchmark video source, which comprises over 30 million 2-minute video clips. 
We apply our video accumulation stage in \autoref{sec:action_video_accu} to obtain some video clips, and manually check these clips to obtain 1,140 video clips.
Then, human annotators are required to describe the number of human subjects, noting static attributes (gender, age, clothing, accessories) for subject identification, followed by body movements, action sequences, and interactions with others in chronological order.

To ensure the quality of annotated captions, we train all annotators for one week. Besides, each video is first annotated by 1 annotator and then checked and made up missing points by another 3 ones. We do not follow previous works like VATEX \cite{wang_vatex_2019} or MSR-VTT \cite{xu_msr-vtt_2016} to annotate multiple captions for one video, since our caption format is an exhaustive description for human actions in a video.

\noindent \textbf{QA Pair Generation.} Based on captions above, we prompt GPT-4o \cite{hurst2024gpt4o} to generate multiple-choice question-answer pairs about human interactions, action details, action sequences, count, and human attributes. The prompts are presented in \autoref{sec_appe:qa_pair_generation}.
Each question-answer pair is checked by 2 annotators and will be corrected if there are any mistakes. All options are shuffled to avoid potential bias.

We get a total of 1,140 video-caption pairs and 7,548 QA pairs in total. Detailed statistics are presented in \autoref{fig:bench_stats}. Although each video clip focuses on a single scene and is relatively short (less than 20 seconds), the captions are highly detailed, often exceeding 100 words. The word cloud analysis reveals that our captions provide comprehensive descriptions. 

For evaluation, we construct the HAICBench-test (refered as HAICBench for short) by selecting 412 videos and 2,000 QA pairs, evenly distributed across all categories. Both HAICBench and the whole QA dataset will be released.

\section{Experiments}

\begin{table*}[t]

% \small

\resizebox{\textwidth}{!}{
\centering

\begin{tabular}{lccccccc}
\hline
\textbf{Model} &
  \textbf{Frames} &
  \textbf{Avg.} &
  \textbf{\begin{tabular}[c]{@{}c@{}}Action\\ Details\end{tabular}} &
  \textbf{\begin{tabular}[c]{@{}c@{}}Action\\ Sequence\end{tabular}} &
  \textbf{Interaction} &
  \textbf{Count} &
  \textbf{Attribute} \\ \hline
 Human Annotated Captions  & -  & 96.7          & 100.0          & 94.3          & 95.0          & 95.3          & 99.0          \\
\hline
Gemini-1.5-Pro \cite{team2024gemini}         & -  & 41.0 & 33.7 & 30.4 & 36.0 & 46.0 & 58.9          \\
GPT-4o \cite{hurst2024gpt4o}                 & 50*  & 40.0 & 37.8 & 30.6 & 35.8 & 44.6 & 51.4            \\ \hline
VideoLLaMA2-7B \cite{cheng_videollama2_2024} & 64 & 21.0 & 15.7 & 18.9 & 17.3 & 28.9 & 24.2          \\
LongVA-7B \cite{zhang_longva_2024}           & 64 & 19.3 & 14.7 & 13.8 & 11.5 & 36.0 & 20.3          \\
InternVL2-8B \cite{chen_internvl2_2024}      & 64 & 28.8 & \textbf{31.0} & 19.9 & 21.5 & 31.5 & 40.2          \\
Qwen2VL-7B \cite{wang_qwen2vl_2024}          & 64 & 27.0 & 18.5 & \textbf{22.1} & 19.9 & 35.7 & 38.9          \\
LLaVA-Video-7B \cite{zhang_llava-video_2024} & 64 & 29.2 & 23.1 & 15.8 & 23.7 & 35.0 & 48.3          \\ \hline
LLaVA-Video-ActionPro-7B                     & 64 & \textbf{35.2} & 27.5 & 16.9 & \textbf{30.7} & \textbf{44.2} & \textbf{56.9} \\ \hline
\end{tabular}

}

\caption{Results on HAICBench in the caption evaluation setting. We adopted MLLMs to generate video captions, based on which we prompted an LLM to answer video questions.  *The GPT-4o API supports a maximum of 50 frames per video.}
\label{tab:caption_eval_results}
\end{table*}

\begin{table*}[t]
% \small
\centering
\resizebox{\textwidth}{!}{

\begin{tabular}{lccccccc}
\hline
\textbf{Model} &
  \textbf{Frames} &
  \textbf{Avg.} &
  \textbf{\begin{tabular}[c]{@{}c@{}}Action\\ Details\end{tabular}} &
  \textbf{\begin{tabular}[c]{@{}c@{}}Action\\ Sequence\end{tabular}} &
  \textbf{Interaction} &
  \textbf{Count} &
  \textbf{Attribute} \\ \hline
Gemini-1.5-Pro \cite{team2024gemini}         & -  & 67.0 & 71.3 & 52.1 & 73.5 & 56.7 & 81.4          \\
GPT-4o \cite{hurst2024gpt4o}                 & 50*  & 61.2 & 69.9 & 46.4 & 61.6 & 46.3 & 81.8              \\ \hline
VideoLLaMA2-7B \cite{cheng_videollama2_2024} & 64 & 51.4 & 48.5 & 38.8 & 56.3 & 47.0 & 66.3          \\
LongVA-7B \cite{zhang_longva_2024}           & 64 & 56.9 & 55.6 & 41.8 & 58.4 & 52.1 & 76.4          \\
InternVL2-8B \cite{chen_internvl2_2024}      & 64 & 58.8 & 64.2 & 43.0 & 59.7 & 50.0 & 76.9          \\
Qwen2VL-7B \cite{wang_qwen2vl_2024}          & 64 & 62.5 & \textbf{69.4} & 44.7 & 60.4 & 57.5 & 80.6          \\
LLaVA-Video-7B \cite{zhang_llava-video_2024} & 64 & 62.0 & 64.3 & 43.0 & 62.5 & 56.8 & 83.5 \\ \hline
LLaVA-Video-ActionPro-7B                     & 64 & \textbf{64.2} & 65.7 & \textbf{45.0} & \textbf{63.6} & \textbf{61.3} & \textbf{85.3}          \\ \hline
\end{tabular}

}
\caption{Results on HAICBench in the standard evaluation setting. *The GPT-4o API supports a maximum of 50 frames per video.}
\label{tab:standard_eval_results}
\end{table*}

% HAIC v1 data
% Gemini-1.5-Pro \cite{team2024gemini}         & -  & 66.9          & 60.3          & 49.3          & 75.7          & 66.7          & 82.7          \\
% GPT-4o \cite{hurst2024gpt4o}                 & 50*  & 60.6              & 59.1              & 43.4              & 64.0              & 54.0              & 82.3              \\ \hline
% VideoLLaMA2-7B \cite{cheng_videollama2_2024} & 64 & 49.7          & 35.4          & 37.7          & 48.3          & 62.7          & 64.3          \\
% LongVA-7B \cite{zhang_longva_2024}           & 64 & 50.3          & 43.1          & 39.3          & 56.0          & 37.3          & 75.7          \\
% InternVL2-8B \cite{chen_internvl2_2024}      & 64 & 58.2          & 41.4          & 40.0          & 61.7          & 68.0          & 80.0          \\
% Qwen2VL-7B \cite{wang_qwen2vl_2024}          & 64 & 64.4          & \textbf{61.1} & 44.7          & 65.3          & 71.3          & 79.7          \\
% LLaVA-Video-7B \cite{zhang_llava-video_2024} & 64 & 64.3          & 53.7          & 43.7          & 67.3          & 70.7          & \textbf{86.3} \\ \hline
% LLaVA-Video-ActionPro-7B                     & 64 & \textbf{66.4} & 56.6          & \textbf{45.0} & \textbf{68.7} & \textbf{76.0} & 86.0          \\ \hline

\begin{table*}[t]
\small
\centering

\resizebox{\textwidth}{!}{
\begin{tabular}{ccccccccccc}
\hline
\multirow{3}{*}{\textbf{Base Model}} & \multicolumn{6}{c}{\textbf{Training Dataset}} & \multirow{3}{*}{\textbf{MVB}} & \multirow{3}{*}{\textbf{PerTest}} & \multirow{3}{*}{\textbf{ANetQA}} & \multirow{3}{*}{\textbf{HAICBench}} \\ \cline{2-7}
 & \multicolumn{2}{c}{LLaVA-Video-178K} &  & HAICTrain &  & WebVid &  &  &  &  \\ \cline{2-3} \cline{5-5} \cline{7-7}
 & 200K & 126K &  & 126K &  & 126K &  &  &  &  \\ \hline
\multirow{4}{*}{LLaVA-Video} &  &  &  &  &  &  & 59.4 & 58.1 & 61.8 & 62.0 \\
 & \checkmark & \checkmark &  &  &  &  & 60.0 & 58.3 & 63.8 & 63.1 \\
 & \checkmark &  &  & \checkmark &  &  & \textbf{62.1} & \textbf{59.9} & \textbf{65.2} & \textbf{64.2} \\
 & \checkmark &  &  &  &  & \checkmark & 58.8 & 55.8 & 62.1 & 60.8 \\ \hline
\multirow{4}{*}{LLaVA-OneVision} &  &  &  &  &  &  & 57.8 & 47.9 & 57.1 & 59.4 \\
 & \checkmark & \checkmark &  &  &  &  & 59.3 & 55.7 & 61.8 & 60.7 \\
 & \checkmark &  &  & \checkmark &  &  & \textbf{60.6} & \textbf{56.9} & \textbf{62.7} & \textbf{61.3} \\
 & \checkmark &  &  &  &  & \checkmark & 58.8 & 55.8 & 60.4 & 59.8 \\ \hline
\end{tabular}
}

\caption{The gain from training with our high-quality human action captions is effective across several benchmarks. MVB, PerTest, and ANetQA denotes MVBench, PerceptionTest, and ActivityNet-QA, respectively.}
\label{tab:ablation}
\end{table*}

% In this section, we present the experimental setup, evaluation results, and analysis of our-xx-Bench. We evaluate N open-source and proprietary MLLMs, highlighting their strengths and limitations in human caption video scenarios. Building on these initial findings, we then conduct additional in-depth analytical experiments to further explore their performance, aiming to facilitate further advancements for MLLMs in enhancing its human action video understanding capabilities.

\subsection{Experimental Settings} \label{sec:exp_settings}

\noindent \textbf{Baselines.} We selected LLaVA-Video-7B \cite{zhang_llava-video_2024} as our baseline due to its minimal inductive bias in model architecture and superior performance. Furthermore, we enhanced its action understanding capability through post-training, resulting in \textbf{LLaVA-Video-ActionPro-7B}.

\noindent \textbf{Training dataset.} 
To preserve the general capability of LLaVA-Video against catastrophic forgotten, we randomly selected 200K instruction pairs from its training set LLaVA-Video-178K \cite{zhang_llava-video_2024} for sample rehearsal \cite{verwimp_rehearsal_2021}. Then, this subset is combined with our HAICTrain to form a training set with 326K instruction pairs in total.

\noindent \textbf{Evaluation dataset.} To evaluate human action understanding ability, we executed experiments on several action-related benchmarks, including MVBench, ActivityNet-QA,  PerceptionTest, and our HAICBench. 
Note that we focused on questions related to human actions in these benchmarks. For MVBench, we performed a comparison of all sub-tasks whose names contain ``Action'', resulting in 7 types: Action Antonym, Action Count, Action Sequence, Action Prediction, Action Localization, Fine-grained Action, and Unexpected Action. For PerceptionTest, we also selected all questions whose tags are related to action. More details are presented in \autoref{sec_appe:eval_datasets}.

% For human action understanding tasks, we evaluate human action comprehension ability on our HAIC-Bench，use video-question-answer style and video-question-answer style respectively. For general public benchmarks that partially encompass human action QA tasks, we assess our model using three public benchmarks, MVBench\cite{li_mvbench_2024} , ActivityNet-QA\cite{yu_activitynet-qa_2019}, and PerceptionTest\cite{patraucean2024perception}. Specifically, for the MVBench, since we do not focus on scenes and objects and for the fair evaluation of human behavior understanding, we perform a comparison of 6 human behavior-related sub-tasks, which are 1) Action Antonym, 2) Action Count, 3) Action Localization, 4) Action Sequence, 5) Fine-grained Action, 6) Unexpected Action, respectively.

\noindent \textbf{Evaluation metrics.} For multiple-choice questions in MVBench, PerceptionTest, and HAICBench, we followed the prompting approach in LLaVA-1.5 \cite{liu_llava15_2024} and used accuracy as the metric. For ActivityNet-QA which features open-ended questions, we followed the evaluation protocol in Video-ChatGPT \cite{video-chargpt}, utilizing GPT-4o-0513 to calculate accuracy. 
In terms of HAICBench, we implemented two evaluation settings: (1) \textbf{Standard Evaluation}, where the video and question are directly input into an MLLM to generate an answer; and (2) \textbf{Caption Evaluation}, where the model generates a caption for each video, and an LLM (Gemini-1.5-Pro) answers questions based on this caption, thereby assessing caption ability based on QA accuracy. To avoid the LLM guessing answers when captions do not mention related contents, we asked Gemini-1.5-Pro to refuse to answer those questions (see \autoref{sec_appe:cap_eval_setting}).

% For caption, we use prompt  \textit{"<video>\nPlease provide a detailed description of the video. If there are people in the video, your description should describe their attributes and actions in chronological order for each person."}. For multiple-choice, we use prompt: \textit{"<video>{question}\nOptions:\nA. {candidate1}\nB. {candidate2}\nC. {candidate3}\nD. {candidate4}\nE. {candidate5}\nAnswer with the option's letter from the given choices directly.\nAnswer: "}. We use accuracy as the evaluation metric for all multiple-choice questions. 

\noindent \textbf{Implementation Details.} We used LLaVA-Video-7B as initial model and fine-tuned it on the training dataset. All parameters were updated during training. The model was fine-tuned for one epoch with a learning rate of 1e-5 for the LLM and 2e-6 for the vision encoder, with 256 batch size. We sampled 64 frames uniformly in both training and evaluation. During generation, we took the greedy search without randomness. Our experiments utilized 128 NVIDIA A800-80GB GPUs.

\subsection{Results on HAICBench} \label{sec:exp_res}

\noindent \textbf{Caption Evaluation Setting.} We follow the caption evaluation setting of HAICBench, where MLLM baselines generate action-related captions for each video, and then the captions and questions are fed into Gemini-1.5-Pro for answers. The results are presented in \autoref{tab:caption_eval_results}.
Our post-trained LLaVA-Video-ActionPro-7B achieves state-of-the-art performance among open-source MLLMs, with a 6\% performance gain. These results show that training with HAICTrain can significantly improve the caption quality of human actions. Furthermore, using human annotated captions as input can achieve very high accuracy, which means there is still considerable room for improvement. Note that the accuracies of some classes are lower than 25\% because we prompted the model not to answer the question if no clues are found in the caption.

\noindent \textbf{Standard Evaluation Setting.} In this setting, videos and questions are directly fed into MLLMs for question answering. The results are presented in \autoref{tab:standard_eval_results}. 
Our post-trained LLaVA-Video-ActionPro-7B outperforms LLaVA-Video by 2.2\%, achieving state-of-the-art performance among open-source MLLMs. On the action-related subtasks, our model surpasses the baseline LLaVA-Video-7B model by 1\%-2\%. These findings highlight the effectiveness of our HAICTrain dataset for enhancing human action understanding. Additionally, Gemini significantly outperforms GPT-4o, which is the reason why we leverage it to generate the captions of our HAICTrain dataset.

\subsection{Effectiveness of Better Captions} \label{sec:ablation}
In the training set, we combine our HAICTrain with 200K instruction pairs from LLaVA-Video-178K \cite{zhang_llava-video_2024} to prevent catastrophic forgotten. To assess the contribution of 126K video-caption pairs in HAICTrain, we replace them with another 126K samples randomly selected from LLaVA-Video-178K and execute post-training on two different MLLMs.
The results are presented in \autoref{tab:ablation}. With the same amount of data, training with HAICTrain consistently yields better improvements across various action question-answering benchmarks, including MVBench, PerceptionTest, ActivityNet-QA and HAICBench. We also try the same videos in HAICTain with their original captions in WebVid and find using these captions can not improve accuracy. All of these demonstrate that HAICTrain is better for the model to understand human actions.

\subsection{Qualitative Analysis} \label{sec:exp_our_benchmark}

\begin{figure*}[t]
    \includegraphics[width=\linewidth]{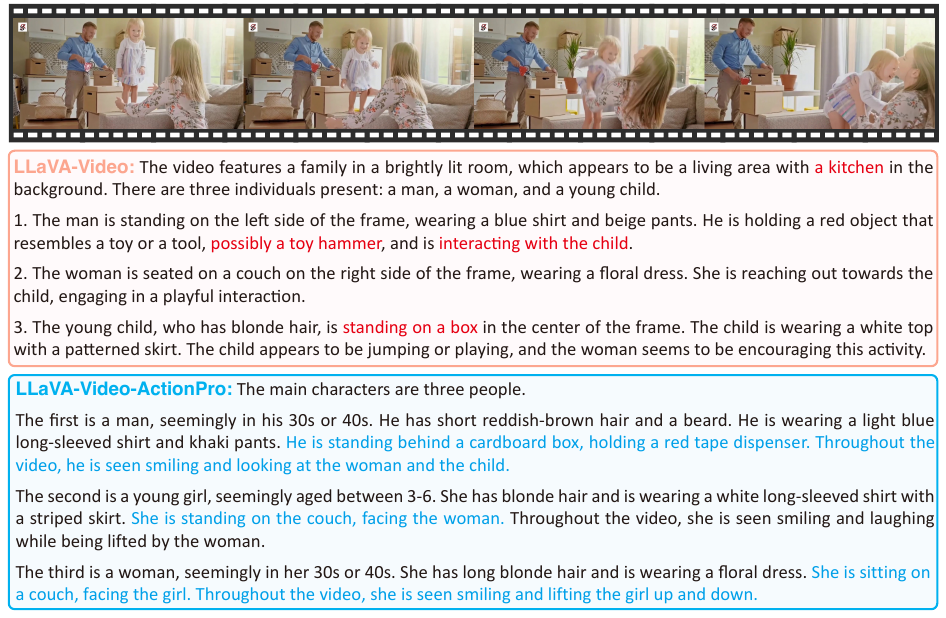}
    \caption{A video caption example in HAICBench.}
    \label{fig:qualitative_analysis}
\end{figure*}

We conduct a qualitative analysis of our post-trained model on HAICBench, as shown in \autoref{fig:qualitative_analysis}. The video features two human boxers fighting on the stage. 
As outlined in the figure, before post-training, the model can only provide a coarse description of their actions. However, after post-training with HAICTrain, our model delivers detailed descriptions of the fighters, including their hair color, clothing, and positions in the cage. Besides, it presents a clear sequence of actions, describing how the attack and defense progress over time. 
This also demonstrates the effectiveness of our caption format. By distinguishing the fighters based on detailed attributes, the caption facilitates better individual recognition, enhancing the viewer's ability to follow the action. 
% More examples can be found in \autoref{sec_appe:qualitative_results}.

\subsection{Effectiveness for Text-video Generation} \label{sec:exp_t2v}

To evaluate our method's effectiveness in text-to-video generation, we use the MovieGenBench dataset \cite{polyak2024moviegenbench}, consisting of 1,003 videos. 
We first take LLaVA-Video-7B and LLaVA-Video-ActionPro-7B to generate captions, which are then fed to HunyuanVideo \cite{hunyuan_video} and Wanx2.1 \cite{tongyi_wanxiang_2023} to generate videos.
Then, five human annotators assess the semantic relevance between videos generated by two captions, classifying them as ``Good'', ``Same'', or ``Bad''. 
The annotators are asked to make decisions upon the original videos in MovieGen.
We measure the GSB score, i.e., the percentage of ``Good'' and ``Same'' relative to that of ``Bad'' and ``Same''.
Results show that LLaVA-Video-ActionPro-7B outperforms LLaVA-Video-7B in generating captions that lead to more semantically accurate videos.
On HunyuanVideo, LLaVA-Video-ActionPro-7B achieves a GSB score of 2.15, and on Wanx2.1, 6.81.
LLaVA-Video-7B often produced verbose but less accurate captions, failing to capture the original video's core actions.
These findings highlight our model's superior ability to comprehend the actions in videos and relay captions that enable high-quality, semantically faithful text-to-video generation.
More cases are presented in Appendix~\ref{sec:t2v_cases}.

\section{Conclusion}

% In this study, we aim to address the challenge of lacking high-quality video-caption data for large video-language models (LVLMs) and text-to-video models (T2VMs). We develop ShareGPT4Video, a high-quality video-caption dataset, and ShareCaptioner-Video, an advanced and versatile model in the video-language multi-modal area. By employing a series of strategies and designs, we generate 40K detailed captions from advanced image multi-modal model, GPT4V, and 4.8M highquality captions from our ShareCaptioner-Video. These captions include rich world knowledge, object attributes, camera movements, and detailed temporal descriptions of events. Our extensive experiments validate the effectiveness of our dataset and captioner in enhancing video understanding and generation tasks. We believe that ShareGPT4Video and ShareCaptioner-Video will serve as essential resources for advancing research in the LVLM and T2VM communities.

This study addresses the challenge of human action understanding by introducing a novel two-stage data annotation pipeline, combining data accumulation, human-machine annotation, and human verification. This process produces two datasets: HAICTrain, which significantly enhances human action understanding and generation, and HAICBench, a comprehensive benchmark. Both datasets will be made open-source to advance research and applications in this field, promoting broader impact across human behavior understanding and generation.

% We introduce a novel data generation pipeline composed of two stages. In the first stage, we carefully design strategies to accumulate videos featuring clear human actions from the Internet. In the second stage, we propose a standardized caption format that leverages human attributes to distinguish different individuals and details human actions and interactions chronologically. 
% With this pipeline, we construct two datasets, namely HAICTrain and HAICBench. HAICTrain contains 126K video-caption pairs in the introduced caption format for training purposes. HAICBench includes 500 video-caption pairs and 1,400 QA pairs, all manually annotated, for comprehensively evaluating human action understanding.
%
%We first introduce a standardized caption format that details human actions and interactions chronologically. We then develop a data generation pipeline that enriches videos with clear actions from the Internet, resulting in HAICBench, a benchmark that evaluates MLLMs on human interaction, action detail, sequence, count, and attributes; and HAICTrain, a training set with 120K video-caption pairs. 
%
% Experimental results confirm the effectiveness of our high-quality captions in enhancing human action captioning and question-answering abilities. Both two datasets will be made open-source to facilitate further research.

\section{Limitations}

Although our work makes notable progress in human action understanding, it still has several limitations. 
Despite their comprehensive nature, our HAIC datasets may not cover the entire spectrum of human actions, particularly those involving complex interactions or cultural nuances. 
Furthermore, our work primarily focuses on visual and textual data, lacking integration with audio data, which could provide additional context for understanding. Future work should aim to incorporate audio and further refine the annotation process to address these limitations.

\section{Acknowledgment}

This work was supported 
in part by the National Natural Science Foundation of China(Grant Nos. 62376069 and 62236003), 
in part by the Young Elite Scientists Sponsorship Program by CAST (Grant No. 2023QNRC001), 
in part by Guangdong Basic and Applied Basic Research Foundation (Grant No. 2024A1515012027), 
in part by Jiangsu Science and Technology Major Program (Grant No. BG2024041), 
and in part by the Shenzhen Science and Technology Program (Grant Nos. KQTD20240729102207002 and ZDSYS20230626091203008).

% Did you report the full text of instructions given to participants, including e.g., screenshots, disclaimers of any risks to participants or annotators, etc.?

\newpage
\bibliography{custom}

\appendix

\section{Comparison with concurrent benchmarks}\label{sec_appe:comp_con_benchs}

We compared our benchmark with three concurrent works on human action understanding---MoVidBench~\cite{chen2024motionllm}, MotionBench~\cite{hong2025motionbench}, and FAVORBench~\cite{tu2025favor}---across three key aspects of human actions:

\begin{itemize}
    \item Action details: Fine-grained motion annotations (e.g., body movements, intensity, direction).
    \item Action transitions: Details on the changes of two actions.
    \item Multi-human interactions: Actions involving multiple humans.
\end{itemize}

Our benchmark is the only one covering all three aspects, as illustrated in \autoref{tab:comp_benchs} below.

\begin{table}[h]
    \centering
    \begin{tabular}{l c c c}
    \hline
    Benchmark & Action \ Details & Action \ Transitions & Multi-Human \ Interactions \\
    \hline
    MoVidBench & $\checkmark$ & $\checkmark$ & None \\
    MotionBench & $\times$ & $\checkmark$ & Few \\
    FAVORBench & $\checkmark$ & $\checkmark$ & Few \\
    HAICBench & $\checkmark$ & $\checkmark$ & Rich \\
    \hline
    \end{tabular}
    \caption{Comparison of recent human action understanding benchmarks on three aspects of human action}
    \label{tab:comp_benchs}
\end{table}

\clearpage

\section{Automatic Annotation} \label{sec_appe:cap_generation}

\noindent\begin{minipage}{\textwidth}
    \centering
    \includegraphics[width=\textwidth]{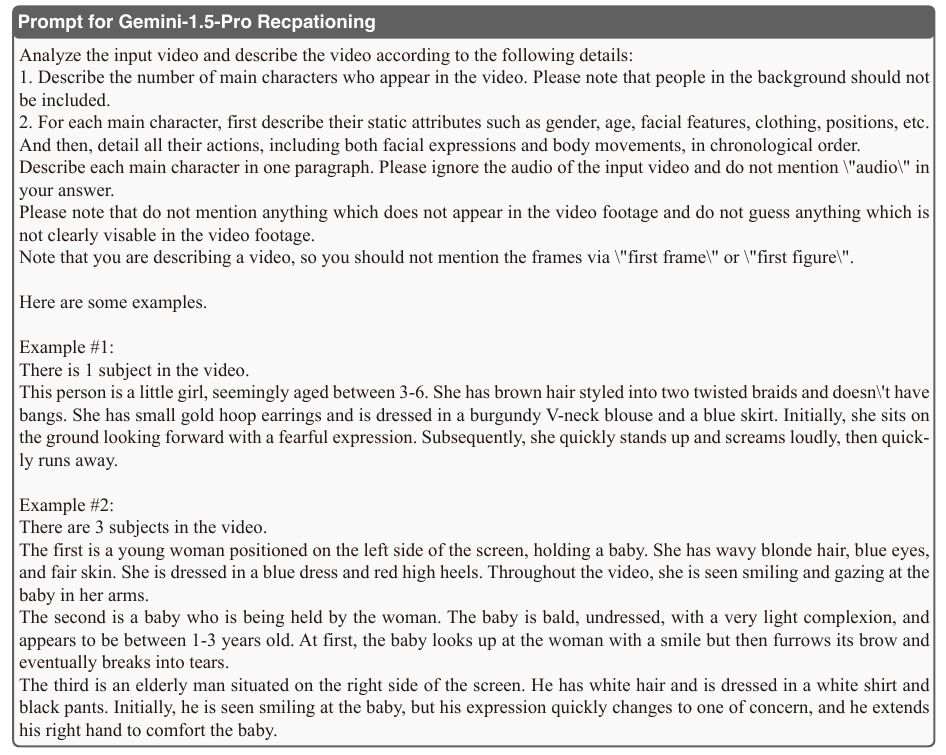}
    \captionof{figure}{Prompt for Gemini-Pro Re-captioning.}
    \label{fig:prompt_gemini_recap}
    \vspace{1em}
\end{minipage}

In \autoref{sec:fluid_cap}, we utilize Gemini-1.5-Pro to generate captions for HAICTrain in our defined standardized format. The prompt is detailed in \autoref{fig:prompt_gemini_recap}.

\section{Evaluation datasets} \label{sec_appe:eval_datasets}

In \autoref{sec:exp_settings}, we use action subsets from MVBench and PerceptionTest as our benchmarks. Here, we explain our subset selection process in detail.

For MVBench, we choose seven categories whose names include "action": \texttt{Action Antonym}, \texttt{Action Count}, \texttt{Action Sequence}, \texttt{Action Prediction}, \texttt{Action Localization}, \texttt{Fine-grained Action}, and \texttt{Unexpected Action}.

For PerceptionTest, we select questions tagged with any of the five labels related to human actions: \texttt{Action counting}, \texttt{Action recognition}, \texttt{Adversarial action}, \texttt{Distractor action}, and \texttt{Occlusion (Occluded interactions)}. \\ \\ \\ \\ \\ \\ \\ \\ \\ \\ \\ \\ \\ \\ \\ \\ \\ \\ \\ \\ \\ \\ \\ \\ \\ \\ \\ \\ \\ \\ \\

\section{QA Pair Generation} \label{sec_appe:qa_pair_generation}

\begin{figure*}[!h]
    \centering
    \includegraphics[width=\textwidth]{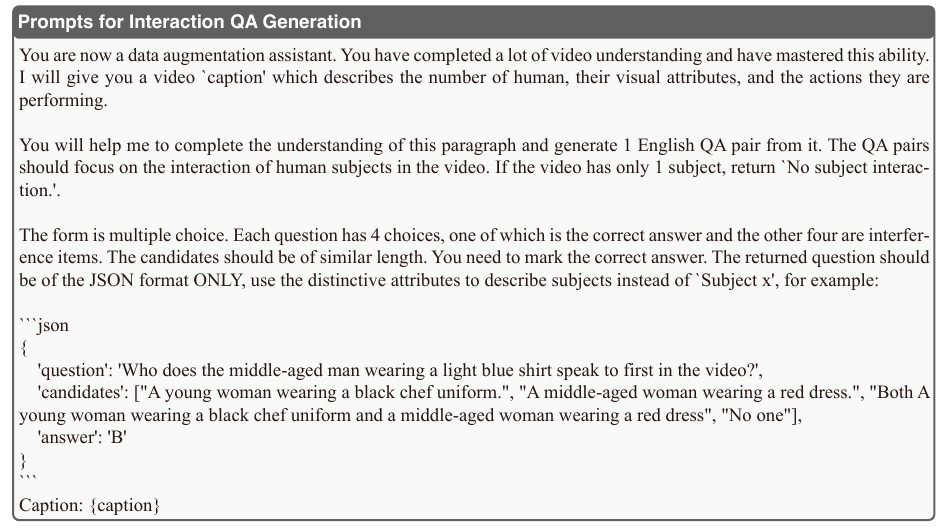}
    \caption{Prompt for action interaction QA generation.}
    \label{fig:prompt_interaction_qa_generation}
\end{figure*}

\begin{figure*}[!h]
    \centering
    \includegraphics[width=\textwidth]{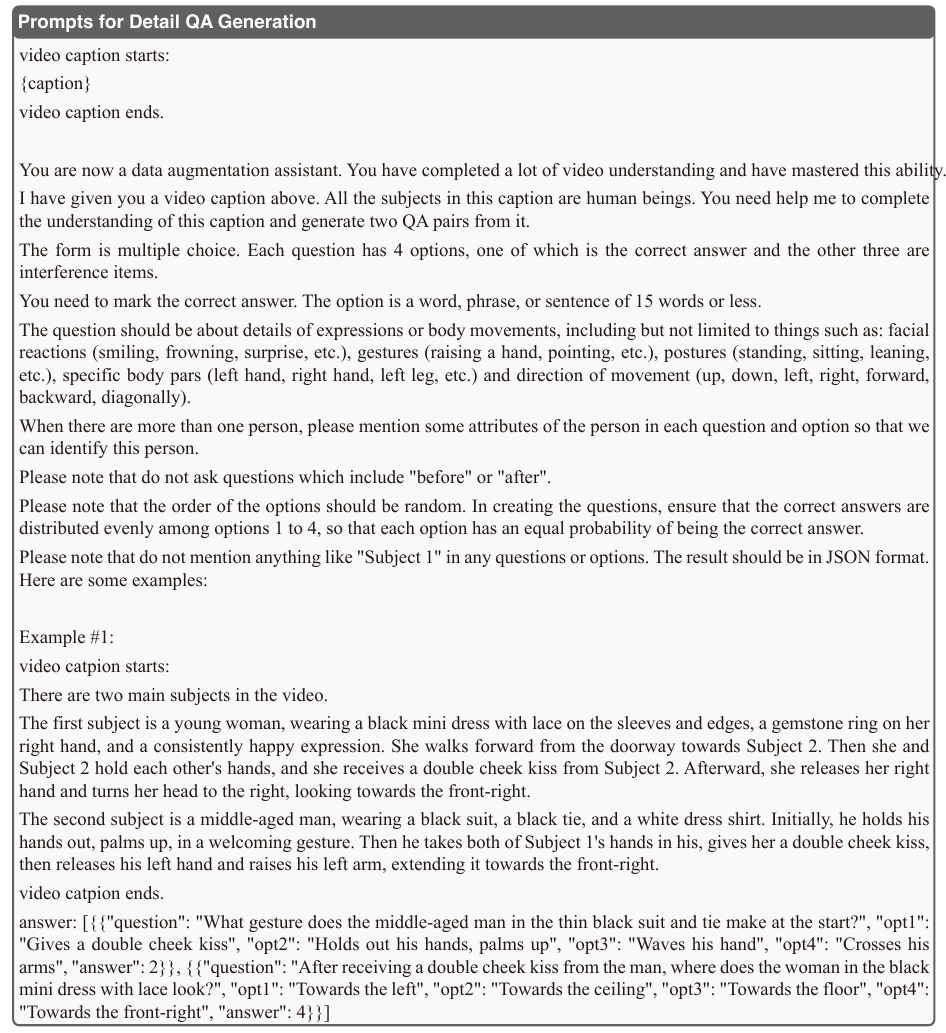}
    \caption{Prompt for action details QA generation.}
    \label{fig:prompt_detail_qa_generation}
\end{figure*}

\begin{figure*}[!h]
    \centering
    \includegraphics[width=\textwidth]{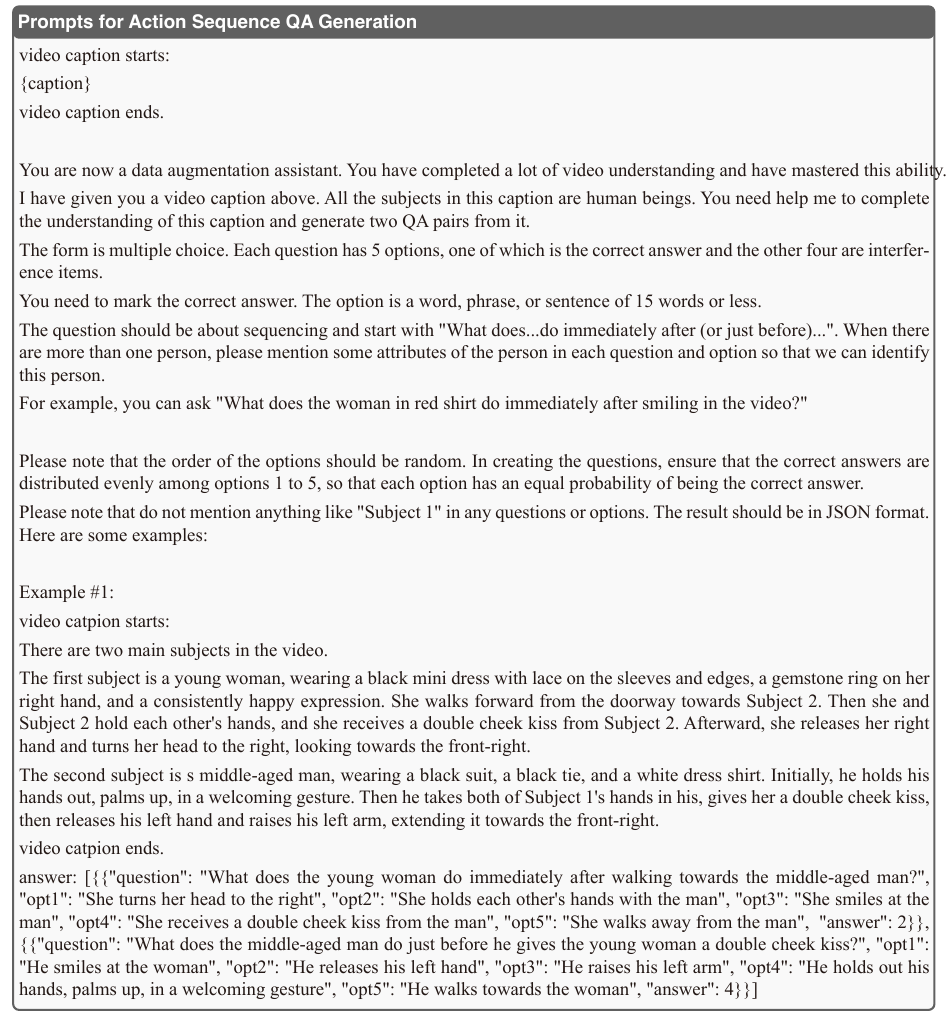}
    \caption{Prompt for action sequence QA generation.}
    \label{fig:prompt_sequence_qa_generation}
\end{figure*}

\begin{figure*}[!h]
    \centering
    \includegraphics[width=\textwidth]{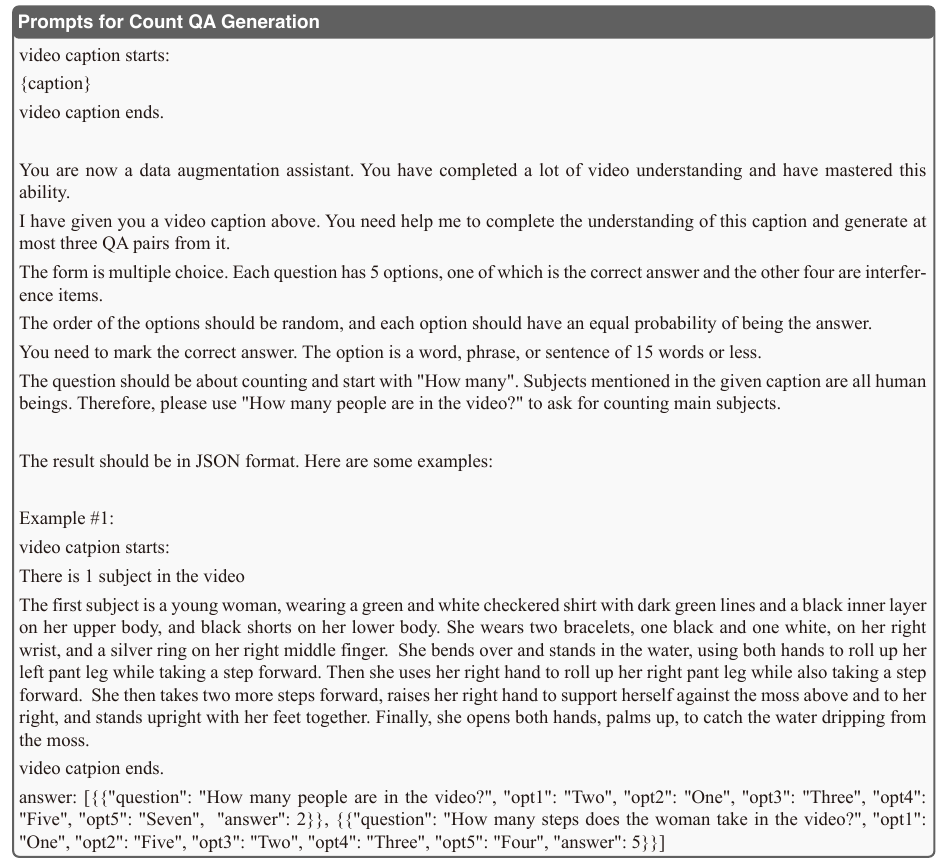}
    \caption{Prompt for action count QA generation.}
    \label{fig:prompt_count_qa_generation}
\end{figure*}

\begin{figure*}[!h]
    \centering
    \includegraphics[width=\textwidth]{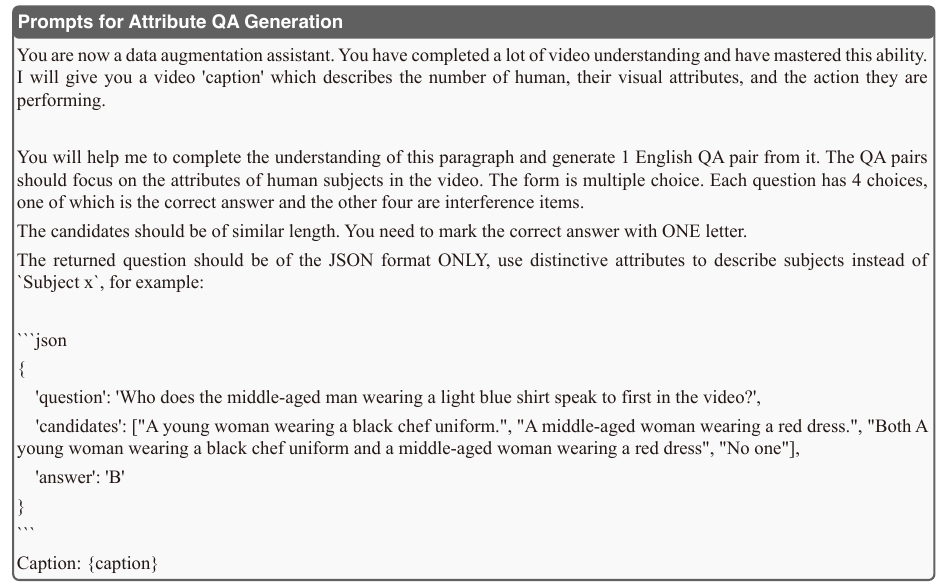}
    \caption{Prompt for human attribute QA generation.}
    \label{fig:prompt_attribute_qa_generation}
\end{figure*}

\begin{figure*}[!t]
    \centering
    \includegraphics[width=\textwidth]{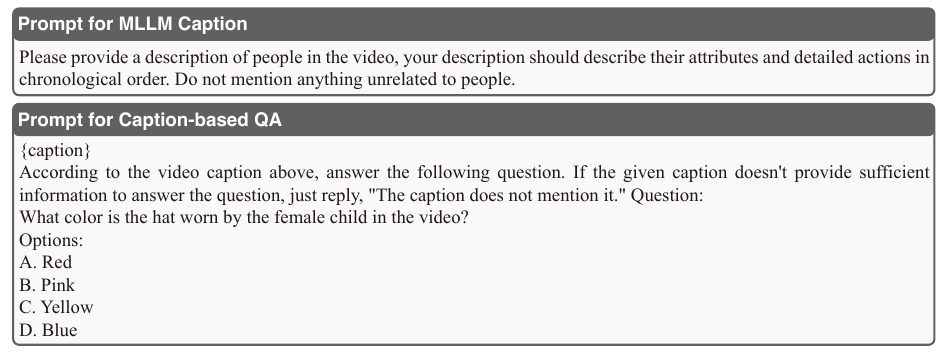}
    \caption{Prompt caption evaluation setting.}
    \label{fig:prompt_caption_based_qa}
\end{figure*}

% \begin{figure*}[!t]
%     \centering
%     \includegraphics[width=\textwidth]{figs/qualitative_analysis_app_1.pdf}
%     \includegraphics[width=\textwidth]{figs/qualitative_analysis_qa.pdf}
%     \caption{Additional qualitative results.}
%     \label{fig:appe_qualitative_analysis_qa}
% \end{figure*}

As briefly discussed in \autoref{sec:fluid_bench}, we utilized a large language model (LLM) to assist in generating QA pairs for HAICBench. The prompts used are as follows: for action interaction QA, see \autoref{fig:prompt_interaction_qa_generation}; for action detail QA, see \autoref{fig:prompt_detail_qa_generation}; for action sequence QA, see \autoref{fig:prompt_sequence_qa_generation}; for action count QA, see \autoref{fig:prompt_count_qa_generation}; and for human attribute QA, see \autoref{fig:prompt_attribute_qa_generation}.

\clearpage

\section{Caption Evaluation Setting} \label{sec_appe:cap_eval_setting}

In \autoref{sec:exp_settings}, we outline our caption evaluation setup, as illustrated in \autoref{fig:prompt_caption_based_qa}. Initially, we prompt MLLMs to generate a caption for each video using a specific prompt. We then combine the generated caption with a question to enable Gemini-1.5-Pro to produce answers.

% \section{Additional Qualitative Results} \label{sec_appe:qualitative_results}

% We provide additional qualitative results in \autoref{fig:appe_qualitative_analysis_qa}. The video features three human subjects: a man packing a box while a child and a woman playing together. Before post-training, the model incorrectly assumes that the man was packing the box with the child. However, after post-training with HAICTrain, our model accurately distinguishes the interactions, correctly identifying that the man in the blue shirt is not interacting with the child or woman.

\begin{figure*}[!t]
    \centering
    \includegraphics[width=0.8\textwidth]{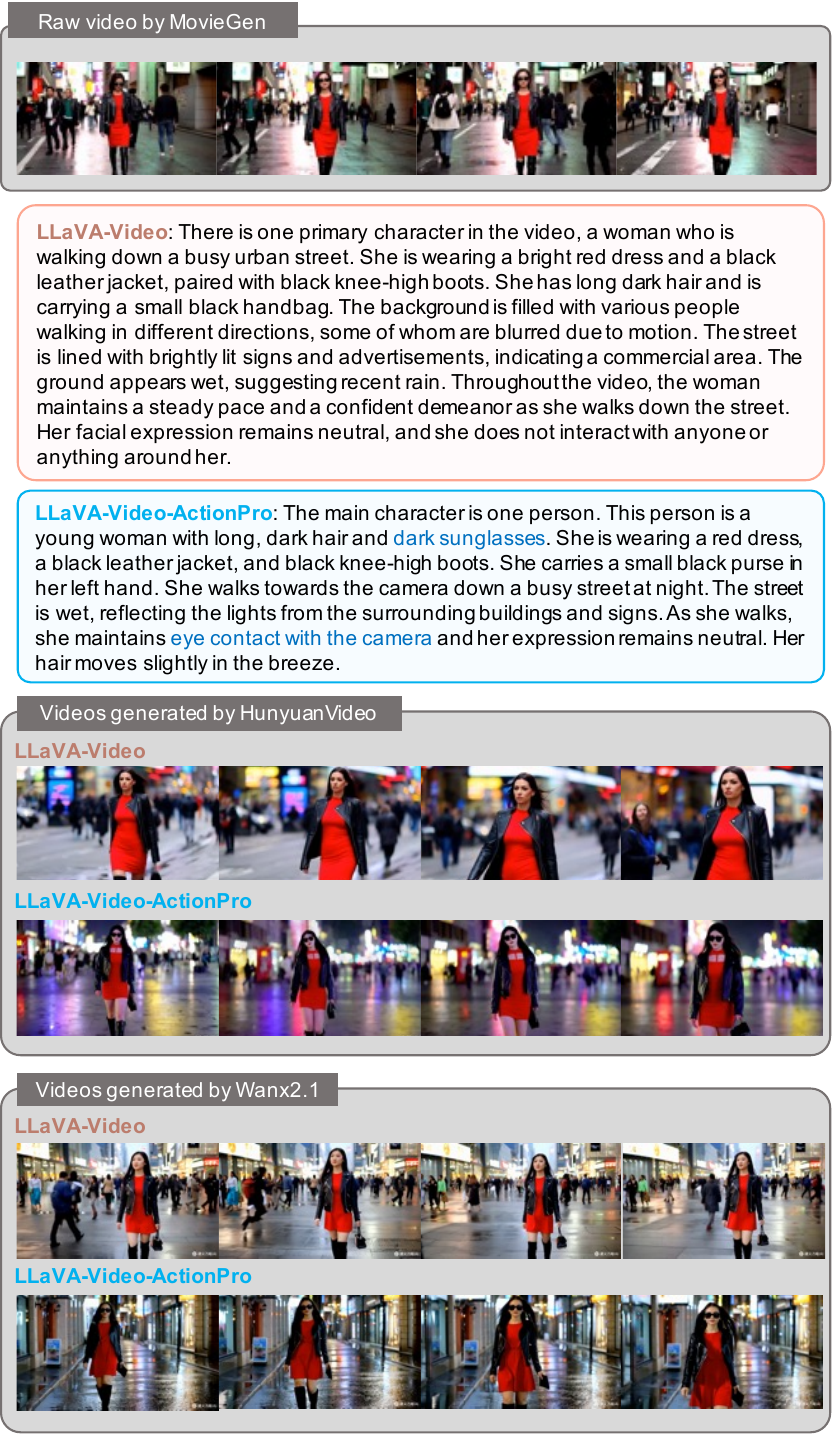}
    \caption{Videos generated by captions from LLaVA-Video and LLaVA-Video-ActionPro of the first sample in MovieGenBench. The main subject in this case is one woman walking along the street. LLaVA-Video-ActionPro provides a more detailed appearance of the woman than LLaVA-Video.}
    \label{fig:t2v_demo_1}
\end{figure*}

\begin{figure*}[!t]
    \centering
    \includegraphics[width=0.8\textwidth]{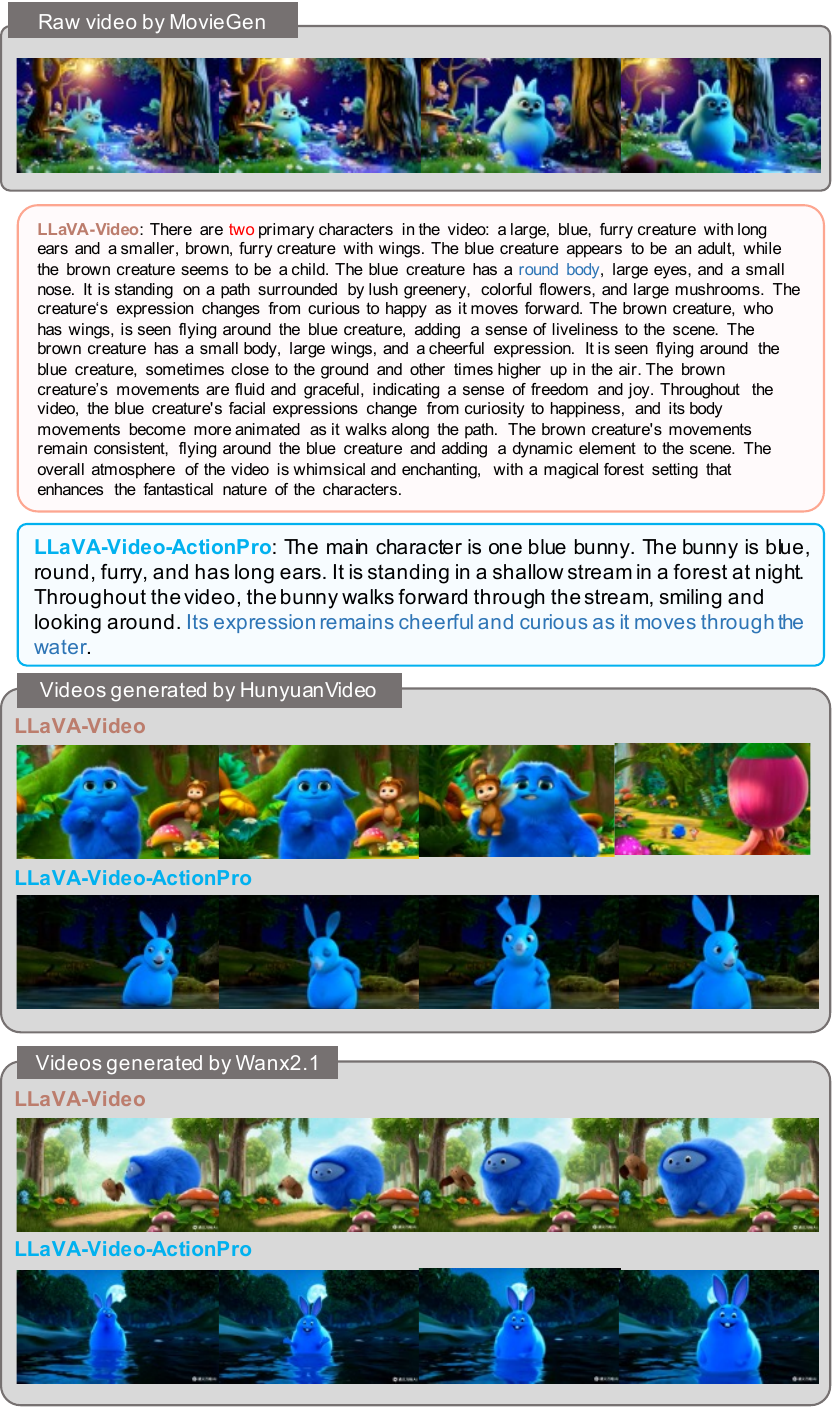}
    \caption{Videos generated by captions from LLaVA-Video and LLaVA-Video-ActionPro of the 17th sample in MovieGenBench. The main subject in this case is one blue animated character. LLaVA-Video incorrectly identifies the main subject.}
    \label{fig:t2v_demo_2}
\end{figure*}

\begin{figure*}[!t]
    \centering
    \includegraphics[width=0.8\textwidth]{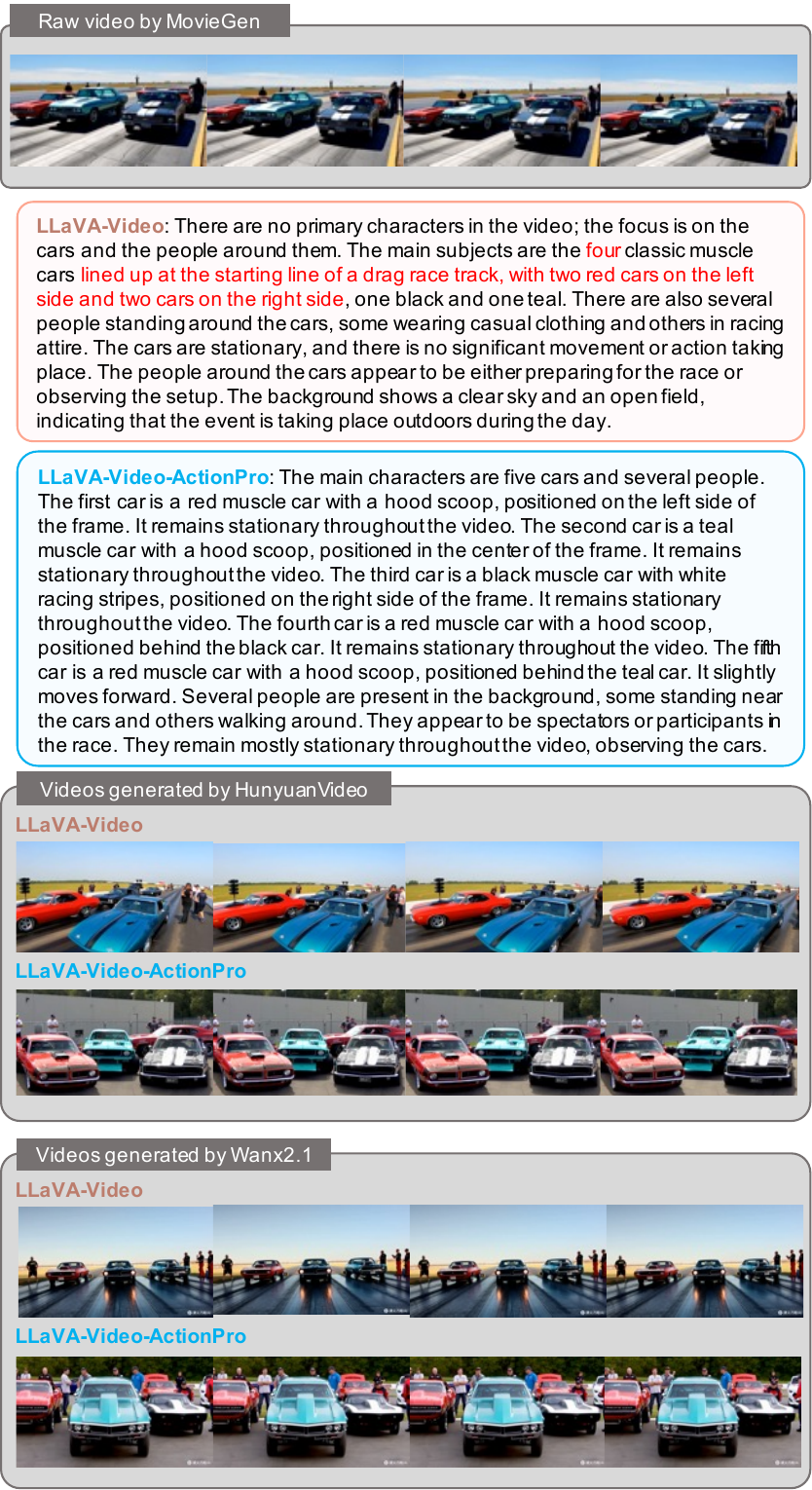}
    \caption{Videos generated by captions from LLaVA-Video and LLaVA-Video-ActionPro of the 129th sample in MovieGenBench. The main subjects in this case are five cars lined in two rows. LLaVA-Video incorrectly identifies the number of the main subjects.}
    \label{fig:t2v_demo_3}
\end{figure*}

\section{Case Study on Text-to-Video Generation} \label{sec:t2v_cases}

We used LLaVA-Video and LLaVA-Video-ActionPro to caption the reference videos (generated by MovieGen) in MovieGenBench.
Then, we used these captions to generate videos, using the open source HunyuanVideo~\footnote{The ``infer-steps'' parameter is set to 50.} and the closed source Wanx2.1~\footnote{https://tongyi.aliyun.com/wanxiang/videoCreation} respectively.
In the manual evaluation, we concluded that the captions generated by LLaVA-Video-ActionPro can give a video that is more consistent and reasonable with the original video.
Figure~\ref{fig:t2v_demo_1} is an example of a single human subject (the first sample in the dataset).
It can be seen that in the caption, LLaVA-Video lost the description of details, resulting in inconsistent generated content.
Figure~\ref{fig:t2v_demo_2} is an example of a single animated character subject (the 17th sample in the dataset).
It can be seen that in the caption, LLaVA-Video incorrectly identified the number of subjects.
Figure~\ref{fig:t2v_demo_3} is an example of multiple subjects (the 129th sample in the dataset).
It can be seen that in the caption, LLaVA-Video incorrectly identified the number of subjects and missed the positional relationship.
Overall, the introduction of refined and formatted action description data has greatly improved the model's ability to understand and retell movements.
This is of great significance for current mainstream action understanding, anomaly recognition, motion generation, and text-to-video generation.

\section{Potential Risks}

The source videos of our HAICBench are from YouTube. YouTube inherently contains social biases, and viewing its videos as representative of 'the world' can perpetuate hegemonic perspectives. The majority of popular YouTubers are men, and the platform's video practices often reflect gender biases. YouTube also faces issues with hate content, including radical alt-right and 'alt-lite' material. These issues are exacerbated by the platform's recommendation algorithm. Even though we downloaded videos independently, filtering them by view count still subjects us to algorithmic influence. The popularity and monetization dynamics on YouTube shape and are shaped by broader cultural trends, affecting the style and content of uploaded videos.

\end{document}